%% file: KimYD_iclr2016.tex
\documentclass{article} 
\usepackage{iclr2016_conference,times}
\usepackage{hyperref}
\usepackage{url}

\usepackage{placeins}
\usepackage{enumitem}
\usepackage{epsfig}
\usepackage{graphicx}
\input{def_boldsymbol.set}

\title{Compression of Deep Convolutional Neural Networks 
for Fast and Low Power Mobile Applications}

\author{Yong-Deok Kim$^1$, Eunhyeok Park$^2$, Sungjoo Yoo$^2$,
Taelim Choi$^1$, Lu Yang$^1$ \& Dongjun Shin$^1$  \\ \\
$^1$Software R\&D Center, Device Solutions, Samsung Electronics, South Korea\\
\texttt{\{yd.mlg.kim, tl.choi, lu2014.yang, d.j.shin\}@samsung.com} \\ \\
$^2$Department of Computer Science and Engineering, Seoul National University, South Korea \\
\texttt{\{canusglow, sungjoo.yoo\}@gmail.com} 
}

\iclrfinalcopy 

\begin{document}

\maketitle

\begin{abstract}
Although the latest high-end smartphone has powerful CPU and GPU,
running deeper convolutional neural networks (CNNs) for complex tasks
such as \emph{ImageNet} classification on mobile devices is challenging.
To deploy deep CNNs on mobile devices, 
we present a simple and effective scheme to compress the entire CNN, 
which we call \emph{one-shot whole network compression}.
The proposed scheme consists of three steps: 
(1) rank selection with variational Bayesian matrix factorization,
(2) Tucker decomposition on kernel tensor, and
(3) fine-tuning to recover accumulated loss of accuracy,
and each step can be easily implemented using publicly available tools.
We demonstrate the effectiveness of the proposed scheme by 
testing the performance of various compressed CNNs 
(\emph{AlexNet}, \emph{VGG-S}, \emph{GoogLeNet}, and \emph{VGG-16})
on the smartphone. 
Significant reductions in model size, runtime,
and energy consumption are obtained, at the cost of small loss in accuracy.
In addition, we address the important implementation level
issue on $1\times1$ convolution, 
which is a key operation of \emph{inception} module of \emph{GoogLeNet} 
as well as CNNs compressed by our proposed scheme. 
\end{abstract}

\section{Introduction}
Deployment of convolutional neural networks (CNNs) 
for computer vision tasks on mobile devices is gaining more and more attention.
On mobile applications, it is typically assumed that 
training is performed on the server and 
test or inference is executed on the mobile devices. 
One of the most critical issues in mobile applications of CNNs is that 
mobile devices have strict constraints in terms of computing power, battery, and memory capacity. 
Thus, it is imperative to obtain CNNs tailored to the limited resources of mobile devices.

Deep neural networks are known to be over-parameterized, 
which facilitates convergence to good local minima of the loss function 
during training \citep{hinton2012,denil2013}.
To improve test-time performance on mobile devices, 
such redundancy can be removed from the trained networks without noticeable impact on accuracy. Recently, there are several studies to apply low-rank approximations to compress CNNs 
by exploiting redundancy \citep{jaderberg2014,denton2014,lebedev2015}. 
Such compressions typically focus on convolution layers since they dominate total computation cost especially in deep neural networks \citep{simonyan2015,szegedy2015}.
Existing methods, though effective in reducing 
the computation cost of a single convolutional layer, 
introduce a new challenge called whole network compression 
which aims at compressing the entire network.

\textbf{Whole network compression:}
It is nontrivial to compress whole and very deep CNNs 
for complex tasks such as \emph{ImageNet} classification.
Recently, \cite{zhang2015a,zhang2015b} showed that
entire convolutional layers can be accelerated with ``asymmetric (3d)" decomposition.
In addition, they also presented the effective rank selection and optimization method.
Although their proposed decomposition of layers can be easily 
implemented in popular development tools (e.g. Caffe, Torch, and Theano),
the rank selection and optimization parts still require
because they consist of multiple steps and depend on the output of previous layers.
In this paper, we present much simpler but still powerful whole network compression scheme
which takes entire convolutional and fully-connected layers into account.

\textbf{Contribution:}
This paper makes the following major contributions.

\begin{itemize}
\item We propose a \emph{one-shot whole network compression scheme} 
which consists of simple three steps: 
(1) rank selection, (2) low-rank tensor decomposition, and (3) fine-tuning.
\item In the proposed scheme, 
Tucker decomposition \citep{tucker1966} with the rank determined 
by a global analytic solution of variational Bayesian matrix factorization (VBMF) \citep{nakajima2012}
is applied on each kernel tensor.
Note that we simply minimize the reconstruction error of linear kernel tensors
instead of non-linear responses. 
Under the Tucker decomposition, the accumulated loss of accuracy 
can be sufficiently recovered by using fine-tuning with \emph{ImageNet} training dataset. 
\item Each step of our scheme can be easily implemented using publicly available tools, 
\citep{nakajima2015} for VBMF, 
\citep{TTB_Software} for Tucker decomposition,
and Caffe for fine-tuning.
\item We evaluate various compressed CNNs 
(\emph{AlexNet}, \emph{VGG-S},\emph{GoogLeNet}, and \emph{VGG-16})
on both Titan X and smartphone.
Significant reduction in model size, runtime,
and energy consumption are obtained, at the cost of small loss in accuracy.
\item By analysing power consumption over time, 
we observe interesting behaviours of $1 \times 1$ convolution
which is the key operation in our compressed model 
as well as in \emph{inception} module of \emph{GoogLeNet}.
Although the $1 \times 1$ convolution is mathematically simple operation,
it is considered to lack in cache efficiency, 
hence it is the root cause of gap between theoretical and practical speed up ratios.
\end{itemize}

This paper is organized as follows. Section 2 reviews related work. Section 3 explains our proposed scheme. Section 4 gives experimental results. Section 5 summarizes the paper.

\section{Related Work}
\subsection{CNN Compression}
CNN usually consists of convolutional layers and fully-connected layers
which dominate computation cost and memory consumption respectively. 
After \cite{denil2013} showed the possibility 
of removing the redundancy of neural networks,
several CNN compression techniques have been proposed.
A recent study \citep{denton2014} showed that the weight matrix of a fully-connected layer 
can be compressed by applying truncated singular value decomposition (SVD) 
without significant drop in the prediction accuracy.
More recently, various methods based on vector quantization \citep{gong2014}, 
hashing techniques \citep{chen2015}, circulant projection \citep{cheng2015},
and tensor train decomposition \citep{novikov2015} were proposed and showed
better compression capability than SVD. 
To speed up the convolutional layers, several methods based on
low-rank decomposition of convolutional kernel tensor 
were proposed \citep{denton2014,jaderberg2014,lebedev2015},
but they compress only single or a few layers.

Concurrent with our work, \cite{zhang2015a} presented ``asymmetric (3d) decomposition"
to accelerate the entire convolutional layers,
where the original $D \times D$ convolution 
is decomposed to $D \times 1$, $1 \times D$, and $1 \times 1$ convolution.
In addition, they also present a rank selection method based on PCA accumulated energy
and an optimization method which minimizes the reconstruction error of non-linear responses.
In the extended version \citep{zhang2015b}, 
the additional fine-tuning of entire network was considered for further improvement.
Compared with these works, our proposed scheme is different in that 
(1) Tucker decomposition is adopted to compress the entire convolutional and fully-connected layers,
(2) the kernel tensor reconstruction error is minimized instead of non-linear response,  
(3) a global analytic solution of VBMF \citep{nakajima2012} is applied 
to determine the rank of each layer, 
and (4) a single run of fine-tuning is performed to account for the accumulation of errors.

A pruning approach \citep{han2015a,han2015b} also aims  at reducing 
the total amount of parameters and operations in the entire network.
Pruning based approaches can give significant reductions in parameter size and computation workload.
However, it is challenging to achieve runtime speed-up with conventional GPU implementation 
as mentioned in \citep{han2015b}.

Orthogonal to model level compression, implementation level approaches were also proposed.
The FFT method was used to speed-up convolution \citep{mathieu2013}.
In \citep{vanhoucke2011}, CPU code optimizations to speed-up 
the execution of CNN are extensively explored.

\subsection{Tensor Decomposition}
\begin{figure}[t]
\centering
\includegraphics[width=0.9\linewidth]{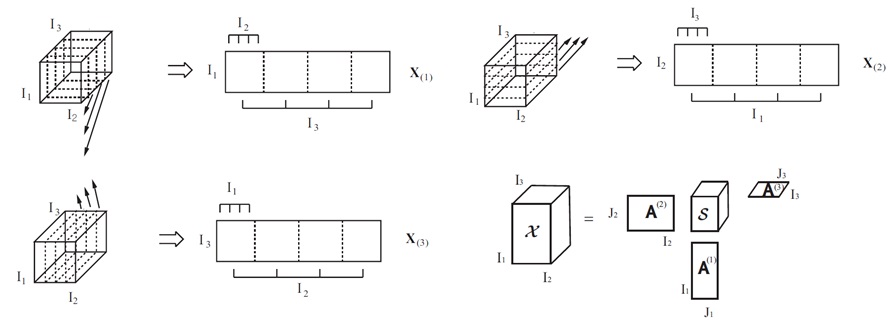}
\caption{Mode-1 (top left), mode-2 (top right), and mode-3 (bottom left) 
matricization of the 3-way tensor. 
They are constructed by concatenation of frontal, horizontal, and vertical slices, respectively.
(Bottom right): Illustration of 3-way Tucker decomposition. 
The original tensor $\calX$ of size $I_1 \times I_2 \times I_3$ is 
decomposed to the product of the core tensor $\calS$ of size 
$J_1 \times J_2 \times J_3$ and factor matrices $\bA^{(1)}$, $\bA^{(2)}$, and $\bA^{(3)}$.}
\label{fig:Tucker}
\vskip -0.1in
\end{figure}

A tensor is a multi-way array of data.
For example, a vector is 1-way tensor and a matrix is 2-way tensor.
Two of the most popular tensor decomposition models are
CANDECOMP/PARAFAC model \citep{carroll1970,harshman1994,shashua2005} and Tucker model \citep{tucker1966,de2000,kim2007}.
In this paper, we extensively use Tucker model for whole network compression.
Tucker decomposition is a higher order extension of 
the singular value decomposition (SVD) of matrix,
in the perspective of computing the orthonormal spaces 
associated with the different modes of a tensor.
It simultaneously analyzes mode-$n$ matricizations 
of the original tensor, and merges them with the core tensor as illustrated in Fig. \ref{fig:Tucker}.

In our whole network compression scheme, we apply Tucker-2 decomposition,
which is also known as  GLRAM \citep{ye2005},
from the second convolutional layer to the first fully connected layers.
For the other layers, we apply Tucker-1 decomposition, which is equivalent to SVD.
For more information on the tensor decomposition, 
the reader is referred to the survey paper \citep{kolda2009}.

\section{Proposed Method}
\begin{figure}[h]
\centering
\includegraphics[width=1.0\linewidth]{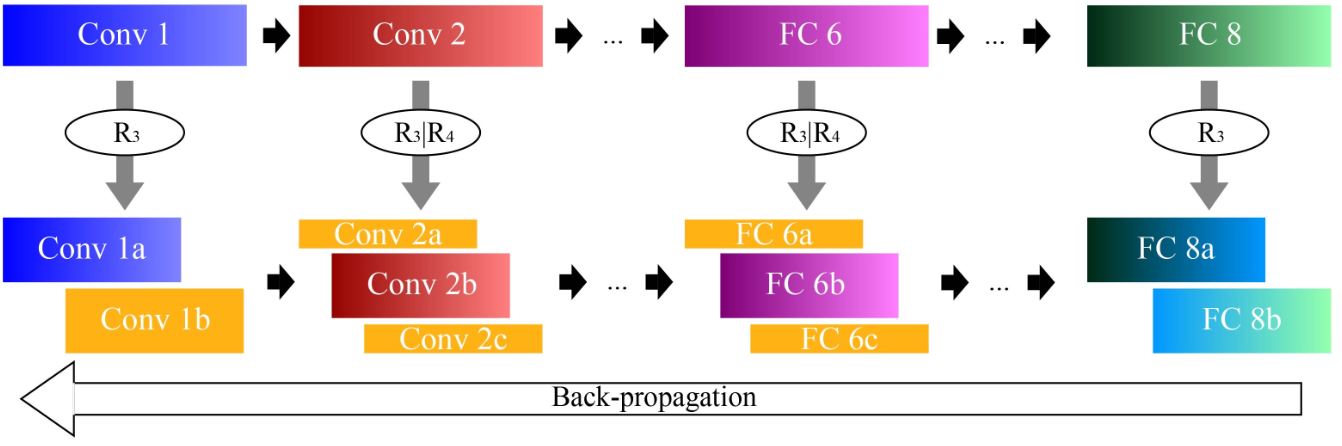}
\caption{Our one-shot whole network compression scheme consists
of (1) rank selection with VBMF; (2) Tucker decomposition on kernel tensor; (3) fine-tuning of entire network. Note that Tucker-2 decomposition is applied from the second convolutional layer
to the first fully connected layers, and Tucker-1 decomposition to the other layers.}
\label{fig:overall}
\end{figure}

Fig. \ref{fig:overall} illustrates our one-shot whole network compression scheme
which consists of three steps: (1) rank selection; (2) Tucker decomposition; (3) fine-tuning. 
In the first step, we analyze principal subspace of mode-3 and mode-4 matricization 
of each layer's kernel tensor with global analytic variational Bayesian matrix factorization.
Then we apply Tucker decomposition on each layer's kernel tensor with previously determined rank.
Finally, we fine-tune the entire network with standard back-propagation.

\subsection{Tucker Decomposition on Kernel Tensor}
\textbf{Convolution kernel tensor:} In CNNs, the convolution operation maps 
an input (source) tensor $\calX$ of size $H \times W \times S$ into
output (target) tensor $\calY$ of size $H' \times W' \times T$ using the following
linear mapping:
\be
\label{eq:kernel_tensor}
\calY_{h',w',t} &=& \sum_{i=1}^D \sum_{j=1}^D \sum_{s=1}^S 
\calK_{i,j,s,t} ~  \calX_{h_i, w_j, s}, \nonumber \\ 
h_i &=&  (h'-1)\Delta + i - P ~~\mbox{and}~~ w_j = (w'-1)\Delta  + j - P, 
\ee
where $\calK$ is a 4-way kernel tensor of size $D \times D \times S \times T$,
$\Delta$ is stride, and $P$ is zero-padding size.

\textbf{Tucker Decomposition:}
The rank-$(R_1,R_2,R_3,R_4)$ Tucker decomposition of 4-way kernel tensor $\calK$ has the form:
\bee
\label{eq:tucker}
\calK_{i,j,s,t} = \sum_{r_1=1}^{R_1} \sum_{r_2=1}^{R_2} \sum_{r_3=1}^{R_3} \sum_{r_4=1}^{R_4}
\calC'_{r_1,r_2,r_3,r_4} U^{(1)}_{i,r_1} U^{(2)}_{j,r_2} U^{(3)}_{s,r_3} U^{(4)}_{t,r_4},
\eee
\noindent
where $\calC'$ is a core tensor of size $R_1 \times R_2 \times R_3 \times R_4$ and
$\bU^{(1)}$, $\bU^{(2)}$, $\bU^{(3)}$, and $\bU^{(4)}$ are factor matrices of sizes
$D \times R_1$, $D \times R_2$, $S \times R_3$, and $T \times R_4$, respectively. 

In the Tucker decomposition, every mode does not have to be decomposed.
For example, we do not decompose mode-1 and mode-2 
which are associated with spatial dimensions because they are already quite small 
($D$ is typically 3 or 5).
Under this variant called Tucker-2 decomposition \citep{tucker1966},
the kernel tensor is decomposed to:
\be
\label{eq:tucker2}
\calK_{i,j,s,t} =  \sum_{r_3=1}^{R_3} \sum_{r_4=1}^{R_4} \calC_{i,j,r_3,r_4} ~ 
U^{(3)}_{s,r_3} ~ U^{(4)}_{t,r_4},
\ee
where $\calC$ is a core tensor of size $D \times D \times R_3 \times R_4$.
After substituting \eqref{eq:tucker2} into \eqref{eq:kernel_tensor},  
performing rearrangements and grouping summands, 
we obtain the following three consecutive expressions
for the approximate evaluation of the convolution \eqref{eq:kernel_tensor}:
\be
\label{eq:conva}
\calZ_{h,w,r_3} &=& \sum_{s=1}^{S} U^{(3)}_{s,r_3} ~ \calX_{h,w,s}, \\
\label{eq:convb}
\calZ'_{h',w',r_4} &=& \sum_{i=1}^D \sum_{j=1}^D \sum_{r_3=1}^{R_3} 
\calC_{i,j,r_3,r_4} ~  \calZ_{h_i, w_j, r_3}, \\
\label{eq:convc}
\calY_{h',w',t} &=& \sum_{r_4=1}^{R_4} U^{(4)}_{t,r_4} ~ \calZ'_{h',w',r_4},
\ee
where $\calZ$ and $\calZ'$ are intermediate tensors of sizes 
$H \times W \times R_3$ and $H' \times W' \times R_4$, respectively.

\begin{figure}[t]
\centering
\includegraphics[width=0.9\linewidth]{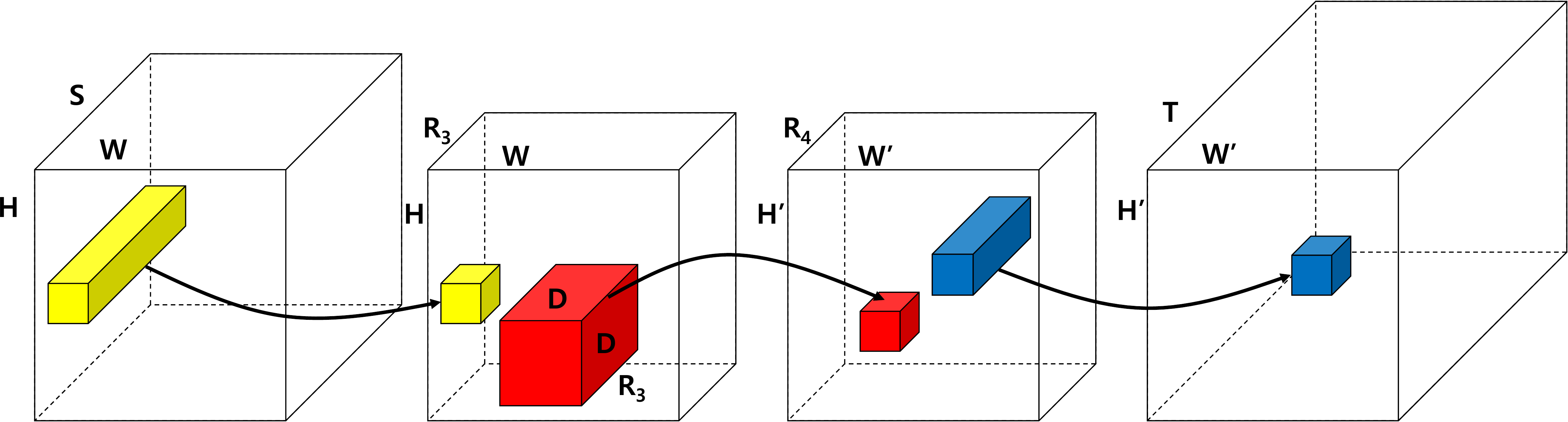}
\caption{Tucker-2 decompositions  for speeding-up a convolution.
Each transparent box corresponds to 3-way tensor $\calX$, $\calZ$, $\calZ'$,
and $\calY$ in (\ref{eq:conva}-\ref{eq:convc}), 
with two frontal sides corresponding to spatial dimensions. 
Arrows represent linear mappings and illustrate how scalar values on the right are computed.
Yellow tube, red box, and blue tube correspond to $1\times 1$, $D \times D$, and $1\times 1$ convolution
in \eqref{eq:conva}, \eqref{eq:convb}, and \eqref{eq:convc} respectively. }
\label{fig:tucker_conv}
\end{figure}

\textbf{$\bone \times \bone$ convolution:}
As illustrated in Fig. \ref{fig:tucker_conv},
computing $\calZ$ from $\calX$ in \eqref{eq:conva} as well as
$\calY$ from $\calZ'$ in \eqref{eq:convc} is $1 \times 1$ convolutions
that essentially perform pixel-wise linear  re-combination of input maps.
It is introduced in \emph{network-in-network} \citep{lin2014} and
extensively used in \emph{inception} module of \emph{GoogLeNet} \citep{szegedy2015}.
Note that computing \eqref{eq:conva} is similar to
inception module in the sense that $D \times D$ convolution is applied after
dimensional reduction with $1 \times 1$ convolution,
but different in the sense that there is no non-linear \emph{ReLU} function
between \eqref{eq:conva} and \eqref{eq:convb}.
In addition, similar to \citep{zhang2015a,zhang2015b},
we compute smaller intermediate output tensor $\calZ'$ in \eqref{eq:convb}
and then recover its size in \eqref{eq:convc}.
The Tucker-2 decomposition naturally integrates two compression techniques.

\textbf{Complexity analysis:} The convolution operation in \eqref{eq:kernel_tensor}
requires $D^2 ST$ parameters and $D^2 ST H'W'$ \emph{multiplication-addition} operations.
With Tucker decomposition, compression ratio $M$ and speed-up ratio $E$ are given by:
\bee
M = \frac{D^2 ST}{S R_3 + D^2 R_3 R_4 + T R_4} ~~~\mbox{and}~~~
E = \frac{D^2 ST H' W'}{S R_3 H W + D^2 R_3 R_4 H' W' + T R_4 H' W'},
\eee
and these are bounded by $S T / R_3 R_4$.

\textbf{Tucker vs CP:}
Recently, CP decomposition is applied to approximate 
the convolution layers of CNNs for \emph{ImageNet} 
which consist of 8 layers \citep{denton2014,lebedev2015}.
However it cannot be applied to the entire layers 
and the instability issue of low-rank CP decomposition is reported \citep{de2008,lebedev2015}.
On the other hand, our kernel tensor approximation with Tucker decomposition
can be successfully applied to the entire layers of 
\emph{AlexNet}, \emph{VGG-S}, \emph{GoogLeNet}, and \emph{VGG-16}

\subsection{Rank Selection with Global Analytic VBMF}
The rank-$(R_3, R_4)$ are very important hyper-parameters
which control the trade-off between 
performance (memory, speed, energy) improvement and accuracy loss.
Instead of selecting the rank-$(R_3, R_4)$ by time consuming trial-and-error,
we considered data-driven one-shot decision 
via \emph{empirical Bayes} \citep{mackay1992}
with \emph{automatic relevance determination} (ARD) prior \citep{tipping2001}.

At the first time, we designed probabilistic Tucker model which is similar to \citep{morup2009},
and applied empirical variational Bayesian learning.
However, the rank selection results were severely  unreliable because
they heavily depend on 
(1) initial condition,
(2) noise variance estimation policy, and
(3) threshold setting for pruning.
For this reason, we decided to use a sub-optimal but highly reproducible approach.

We employed recently developed global analytic solutions for 
variational Bayesian matrix factorization (VBMF) \citep{nakajima2013}.
The global analytic VBMF is a very promising tool 
because it can automatically find noise variance, rank and 
even provide theoretical condition for perfect rank recovery \citep{nakajima2012}.
We determined the rank $R_3$ and $R_4$ by applying global analytic VBMF 
on mode-3 matricization (of size $S \times TD^2$)
and mode-4 matricization (of size $T \times D^2S$) of kernel tensor $\calK$, respectively.

\subsection{Fine-Tuning}
\begin{figure}[t]
\centering
\includegraphics[width=0.6\linewidth]{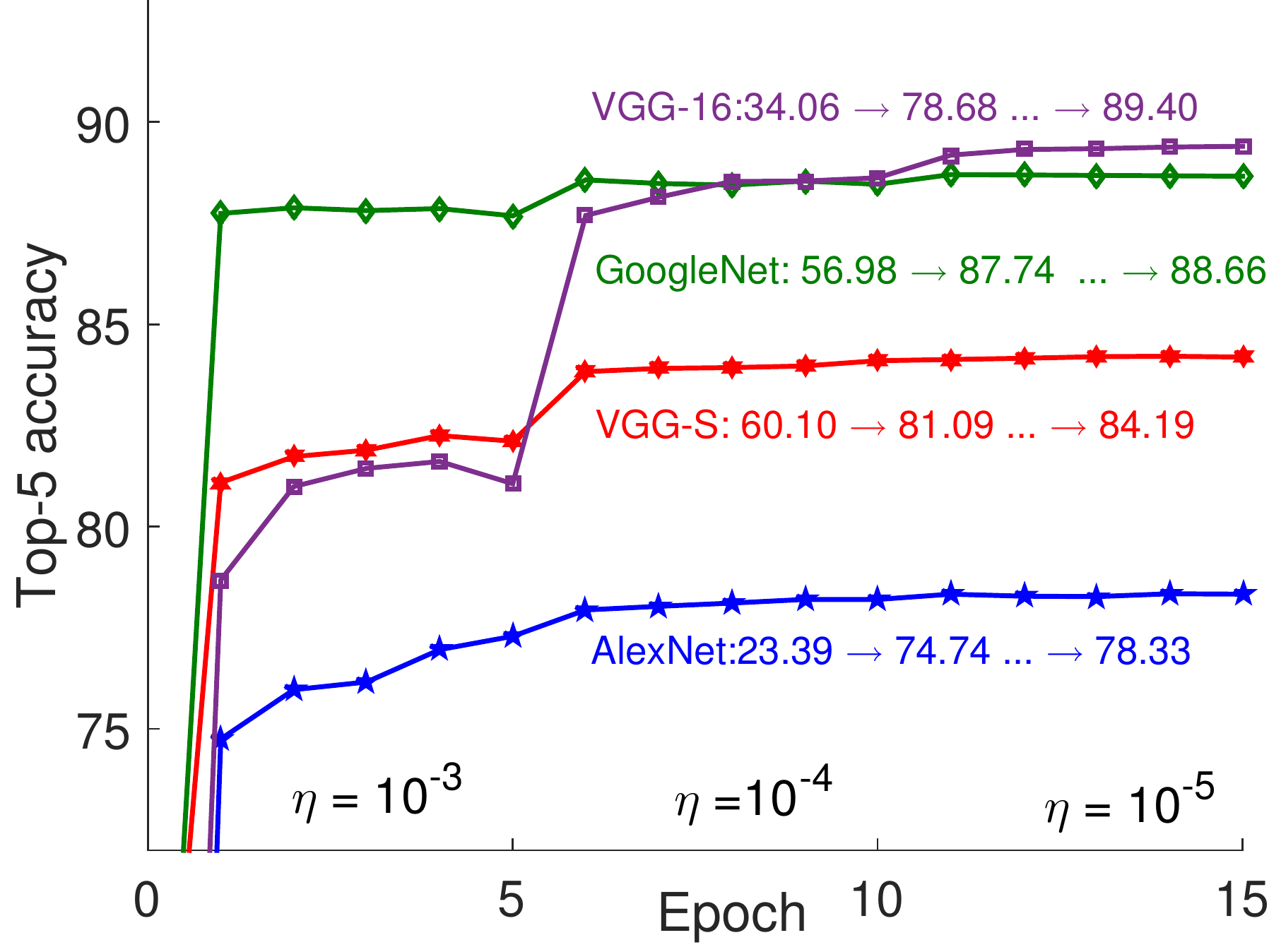}
\caption{Accuracy of compressed CNNs in fine-tuning.}
\label{fig:fine_tuning}
\end{figure}

Because we minimize the reconstruction error of linear kernel tensors instead of non-linear responses, 
the accuracy is significantly dropped after whole network compression
(e.g. more than 50$\%$ in the case of \emph{AlexNet}).
However, as shown in Fig. \ref{fig:fine_tuning}, 
we can easily recover the accuracy by using fine-tuning with \emph{ImageNet} training dataset.
We observed that accuracy is recovered quickly in one epoch. 
However, more than 10 epochs are required to recover the original accuracy.

While \citep{lebedev2015, zhang2015b} reported difficulty on finding a good SGD learning rate,
our single learning rate scheduling rule works well for various compressed CNNs.
In our experiment, we set the base learning $\eta=10^{-3}$ 
and decrease it by a factor of 10 every 5 epochs.
Because of GPU memory limitation, 
we set the batch size: 128, 128, 64, and 32 for 
\emph{AlexNet}, \emph{VGG-S}, \emph{GoogLeNet}, and \emph{VGG-16}, respectively.

We also tried to train the architecture of the approximated model
from scratch on the \emph{ImageNet} training dataset.
At this time, we only tested the Gaussian random initialization and it did not work.
We leave the use of other initialization methods \citep{glorot2010,he2015}
and \emph{batch normalization} \citep{ioffe2015} as future work.

\section{Experiments}
We used four representative CNNs, \emph{AlexNet}, \emph{VGG-S}, \emph{GoogLeNet},
and \emph{VGG-16},\
which can be downloaded on Berkeley's \emph{Caffe model zoo}.
In the case of \emph{inception} module of \emph{GoogLeNet},
we only compressed the $3\times3$ convolution kernel which is the main computational part.
In the case of \emph{VGG-16}, we only compressed the convolutional layers
as done in \citep{zhang2015b}.
Top-5 single-view accuracy is measured using 
50,000 validation images from the \emph{ImageNet2012} dataset.

We performed experiments on Nvidia Titan X (for fine-tuning and runtime comparison on Caffe+cuDNN2) 
and a smartphone, Samsung Galaxy S6 (for the comparison of runtime and energy consumption). 
The application processor of the smartphone (Exynos 7420) is equipped with a mobile GPU, 
ARM Mali T760. 
Compared with the GPU used on Titan X, 
the mobile GPU gives 35 times (6.6TFlops vs 190GFlops) lower computing capability 
and 13 times (336.5GBps vs 25.6GBps) smaller memory bandwidth. 

In order to run Caffe models on the mobile GPU, 
we developed a mobile version of Caffe called S-Caffe (Caffe for Smart mobile devices) 
where all the Caffe models can run on our target mobile devices 
(for the moment, Samsung smartphones) without modification. 
We also developed an Android App which performs image classification 
by running each of the four CNNs (\emph{AlexNet}, \emph{VGG-S}, \emph{GoogLeNet}
, and \emph{VGG-16}) on the smartphone. 

We measured the power consumption of whole smartphone 
which is decomposed into the power consumption of GPU, main memory, 
and the other components of smartphone, 
e.g., ARM CPU, display, modem, etc. and give component-level analysis, 
especially, the power consumption of GPU and main memory
(see supplementary material  for details of measurement environment).
The measurement results of runtime and energy consumption are the average of 50 runs.

\subsection{Overall Results}
Table \ref{tab:overall_results} shows the overall results for the three CNNs. 
Our proposed scheme gives  $\times5.46/\times2.67$ (\emph{AlexNet}), 
$\times7.40/\times4.80$ (\emph{VGG-S}), 
$\times1.28/\times2.06$ (\emph{GoogLeNet}),
and $\times1.09/\times4.93$ (\emph{VGG-16}) reductions 
in total weights and FLOPs, respectively. 
Such reductions offer $\times1.42\sim\times3.68$ ($\times1.23\sim\times2.33$) runtime
improvements on the smartphone (Titan X). 
We report the energy consumption of mobile GPU and main memory.
The smartphone gives larger reduction ratios (e.g., $\times3.41$ vs. $\times2.72$ for \emph{AlexNet}) 
for energy consumption than runtime. 
We will give a detailed analysis in the following subsection.

\textbf{Comparison with \cite{zhang2015b}'s method:}
The accuracy of our compressed \emph{VGG-16} is 89.40$\%$ for theoretical $\times 4.93$ speed-up,
and it is comparable to the 89.6$\%$ (88.9$\%$) for theoretical $\times 4$ ($\times 5$) speed-up
in  \citep{zhang2015b}.
 
\begin{table}[h]
\caption{Original versus compressed CNNs. 
Memory, runtime and energy are significantly reduced with only minor accuracy drop.
We report the time and energy consumption for processing 
single image in S6 and Titan X.
(* compression, S6: Samsung Galaxy S6).}
\label{tab:overall_results}
\begin{center}
\begin{tabular}{l|r|r|r|r|r|r}
Model & Top-5  & Weights & FLOPs & \multicolumn{2}{c|}{S6 } & Titan X \\  \hline 
\emph{AlexNet} & 80.03 & 61M & 725M & 117ms & 245mJ  & 0.54ms \\ 
\emph{AlexNet}* & 78.33 & 11M & 272M & ~~43ms & ~~72mJ & 0.30ms\\ 
(imp.) & (-1.70) & $(\times 5.46$) & $(\times 2.67$) & $(\times 2.72)$ & $(\times 3.41)$ &$(\times 1.81)$\\ \hline 
\emph{VGG-S} & 84.60 & 103M & 2640M & 357ms & 825mJ & 1.86ms \\ 
\emph{VGG-S}* & 84.05 & 14M & 549M & ~~97ms & 193mJ & 0.92ms\\ 
(imp.) & (-0.55) & $(\times 7.40$) &  $(\times 4.80$)& $(\times 3.68)$ & $(\times 4.26)$ & $(\times 2.01)$\\ \hline
\emph{GoogLeNet} & 88.90 & 6.9M & 1566M & 273ms & 473mJ & 1.83ms \\ 
\emph{GoogLeNet}* & 88.66 & 4.7M & 760M & 192ms & 296mJ& 1.48ms\\ 
(imp.) & (-0.24)  & $ (\times 1.28)$ & $(\times 2.06) $ & $(\times 1.42)$ & $(\times 1.60)$ &$(\times 1.23)$\\  
\hline
\emph{VGG-16} & 89.90 & 138M & 15484M & 1926ms & 4757mJ & 10.67ms \\ 
\emph{VGG-16}* & 89.40 & 127M & 3139M & 576ms & 1346mJ& 4.58ms\\ 
(imp.) & (-0.50)  & $ (\times 1.09)$ & $(\times 4.93) $ & $(\times 3.34)$ & $(\times 3.53)$ &$(\times 2.33)$\\  
\end{tabular}
\end{center}
\end{table}

\subsection{Layerwise Analysis}

\input{table_alexnet}
Tables \ref{tab:AlexNet}, \ref{tab:VGG_S}, \ref{tab:GoogLeNet} and \ref{tab:VGG_16}
\footnote{See supplementary material for Tables \ref{tab:VGG_S}, \ref{tab:GoogLeNet} and \ref{tab:VGG_16}}
show the detailed comparisons. 
Each row has two results (the above one for the original uncompressed CNN 
and the other one for the compressed CNN), and improvements. 
For instance, in Table \ref{tab:AlexNet}, the second convolutional layer having 
the input and output channel dimensions of $48 \times 2$ and $128 \times 2$ 
is compressed to give the Tucker-2 ranks of $25 \times 2$ and $59 \times 2$, 
which reduces the amount of weights from $307K$ to $91K$. 
After compression, a layer in the compressed network performs three matrix multiplications. 
We give the details of three matrix multiplications for each of weights, FLOPs, and runtime. 
For instance, on the smartphone (column S6 in Table \ref{tab:AlexNet}), 
the second convolutional layer of compressed \emph{AlexNet} takes 
10.53ms which is decomposed to 0.8ms, 7.43ms and 2.3ms for the three matrix multiplications.

In Tables \ref{tab:AlexNet}, \ref{tab:VGG_S}, \ref{tab:GoogLeNet} and \ref{tab:VGG_16} 
we have two observations.\\
\textbf{Observation 1:} 
Given a compressed network, 
the smartphone tends to give larger performance gain than the Titan X.
It is mainly because the mobile GPU on the smartphone lacks in thread-level parallelism.
It has 24 times less number of threads (2K vs. 48K in terms of maximum number of threads) 
than that in Titan X.
Compression reduces the amount of weights thereby reducing cache conflicts and memory latency.
Due to the small thread-level parallelism, the reduced latency has more impact 
on the performance of threads on the mobile GPU than that on Titan X.

\textbf{Observation 2:} 
Given the same compression rate, 
the smartphone tends to exhibit larger performance gain at fully-connected layers 
than at convolutional layers.
We think it is also due to the reduced cache conflicts enabled 
by network compression as explained above. 
Especially, in the case of fully-connected layers, 
the effect of weight reduction can give more significant impact
because the weights at the fully-connected layers are utilized only once, 
often called dead-on-arrival (DoA) data. 
In terms of cache performance, 
such DoA data are much more harmful than convolution kernel weights (which are reused multiple times). 
Thus, weight reduction at the fully connected layer 
can give more significant impact on cache performance thereby exhibiting 
more performance improvement than in the case of weight reduction at convolutional layers.

\begin{figure}[t]
\begin{center}
\centerline{
\includegraphics[width=1.95in]{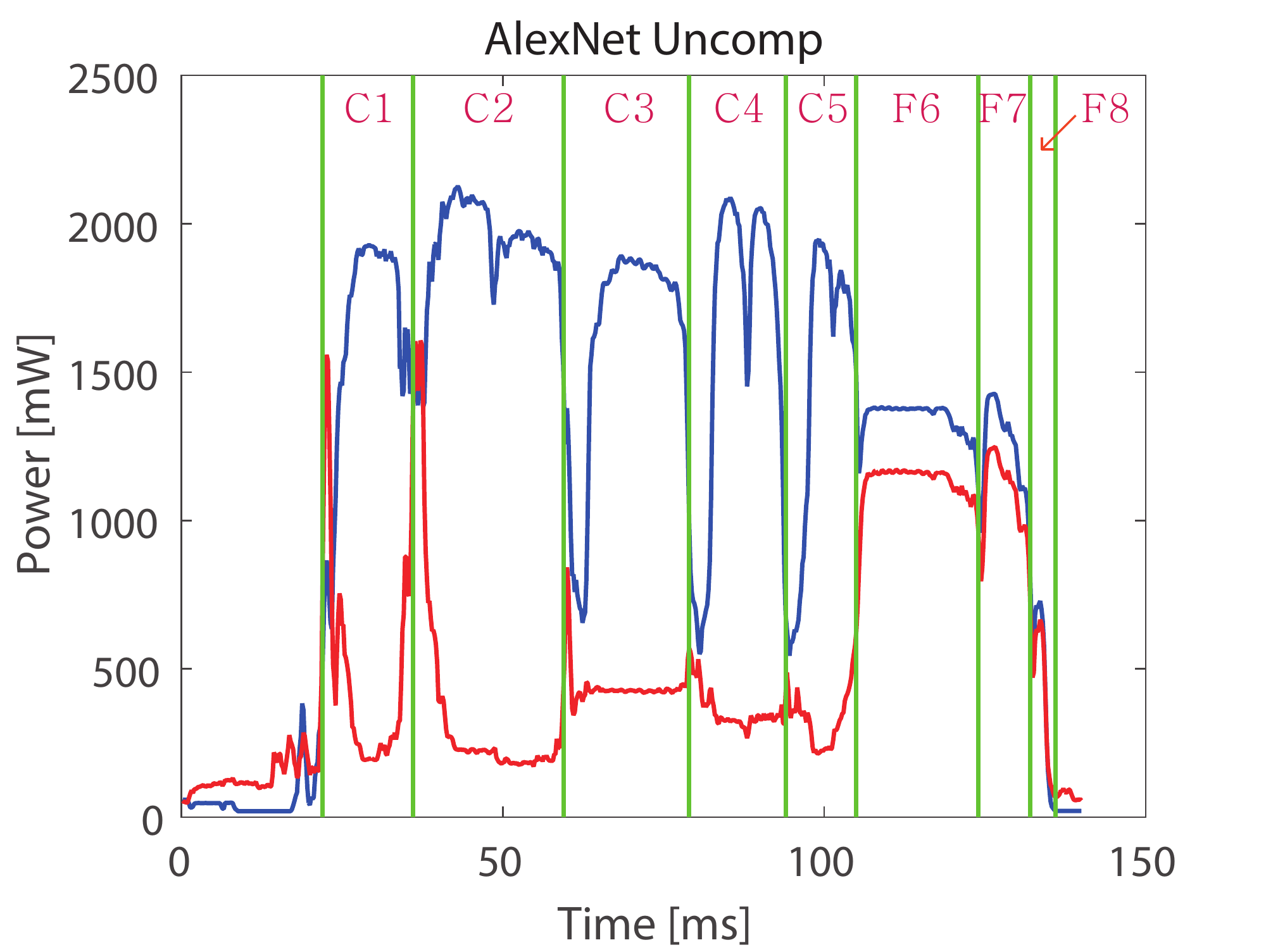}
\includegraphics[width=1.85in]{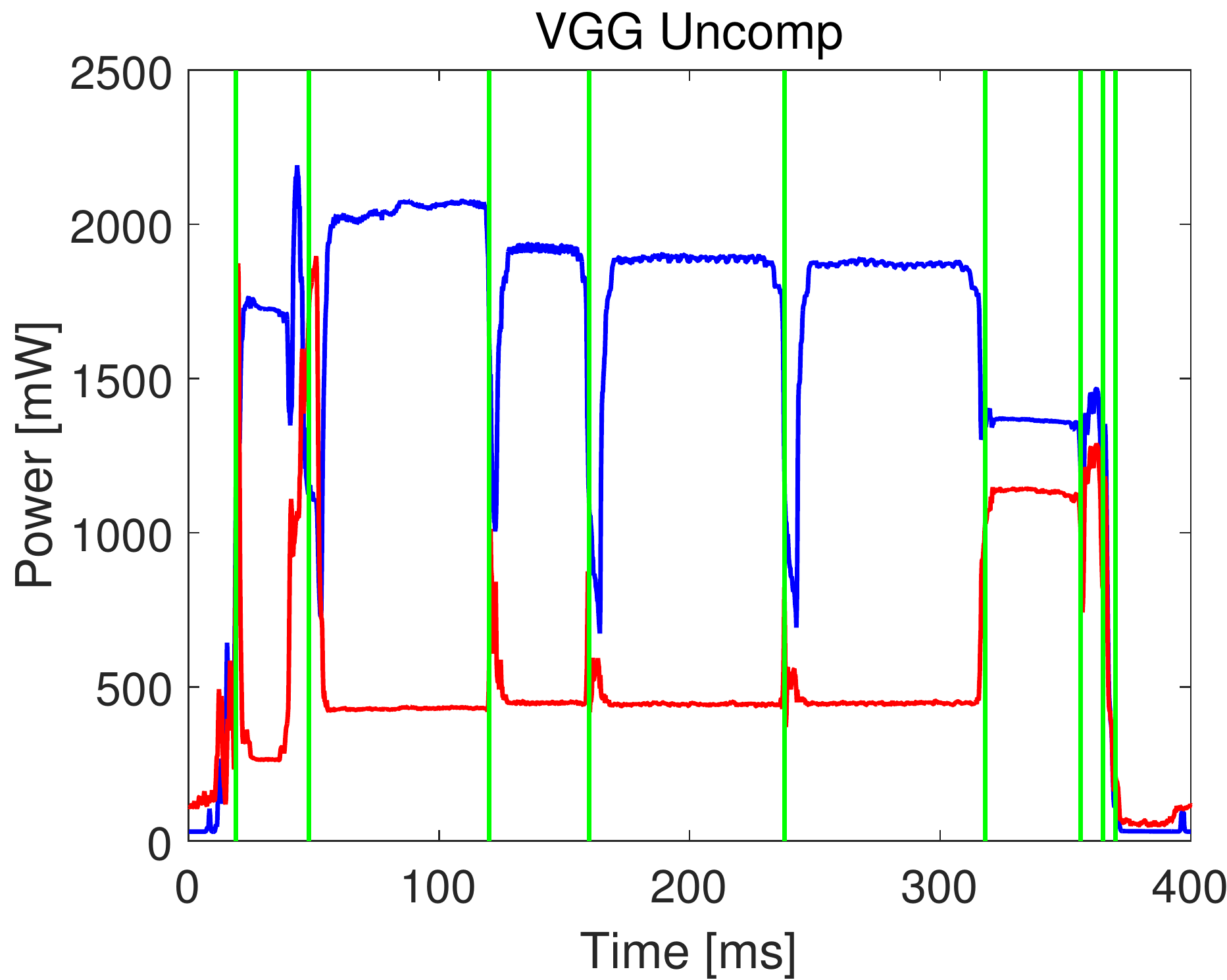}
\includegraphics[width=1.85in]{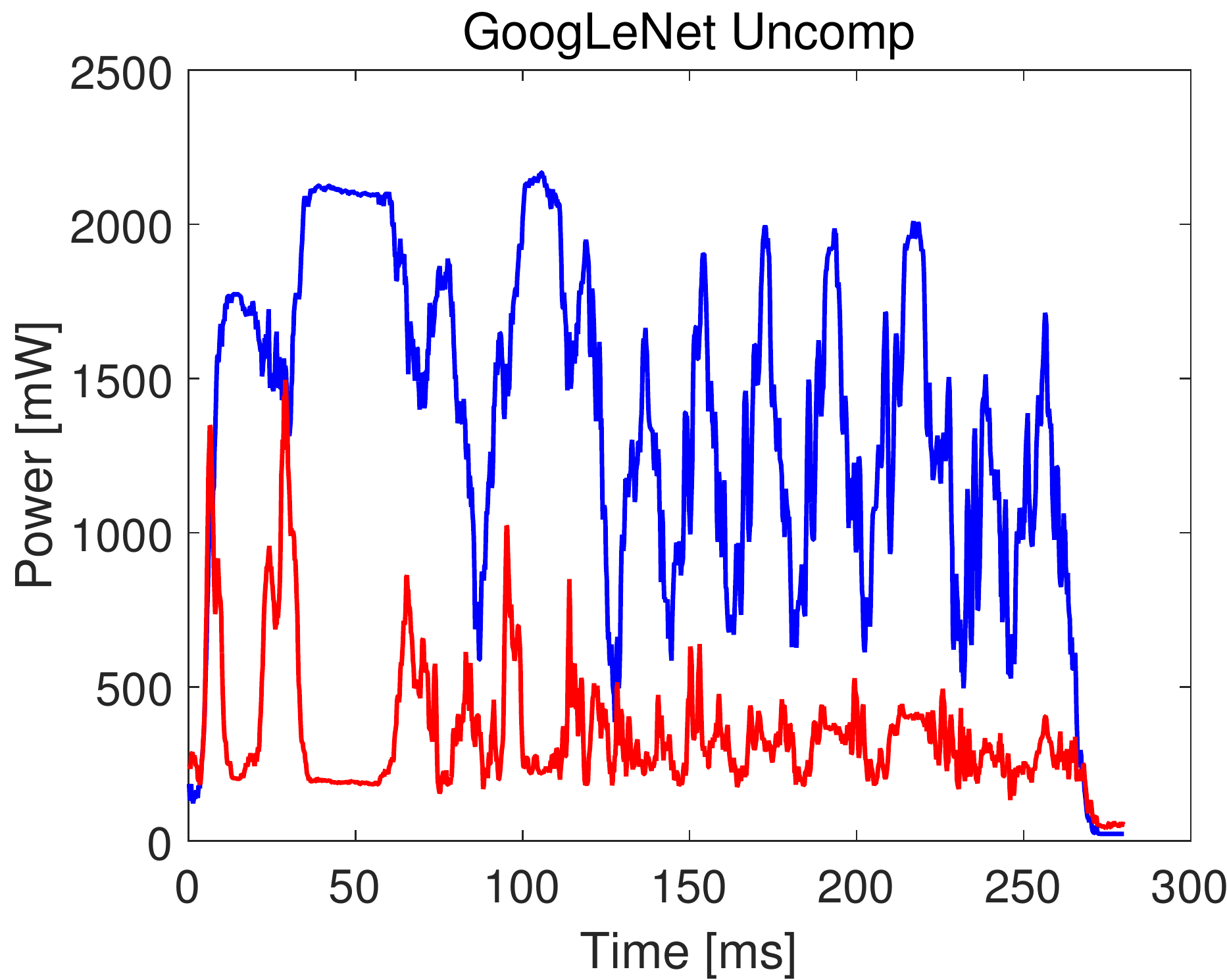}
}
\vspace{0.1in}
\centerline{
\includegraphics[width=1.85in]{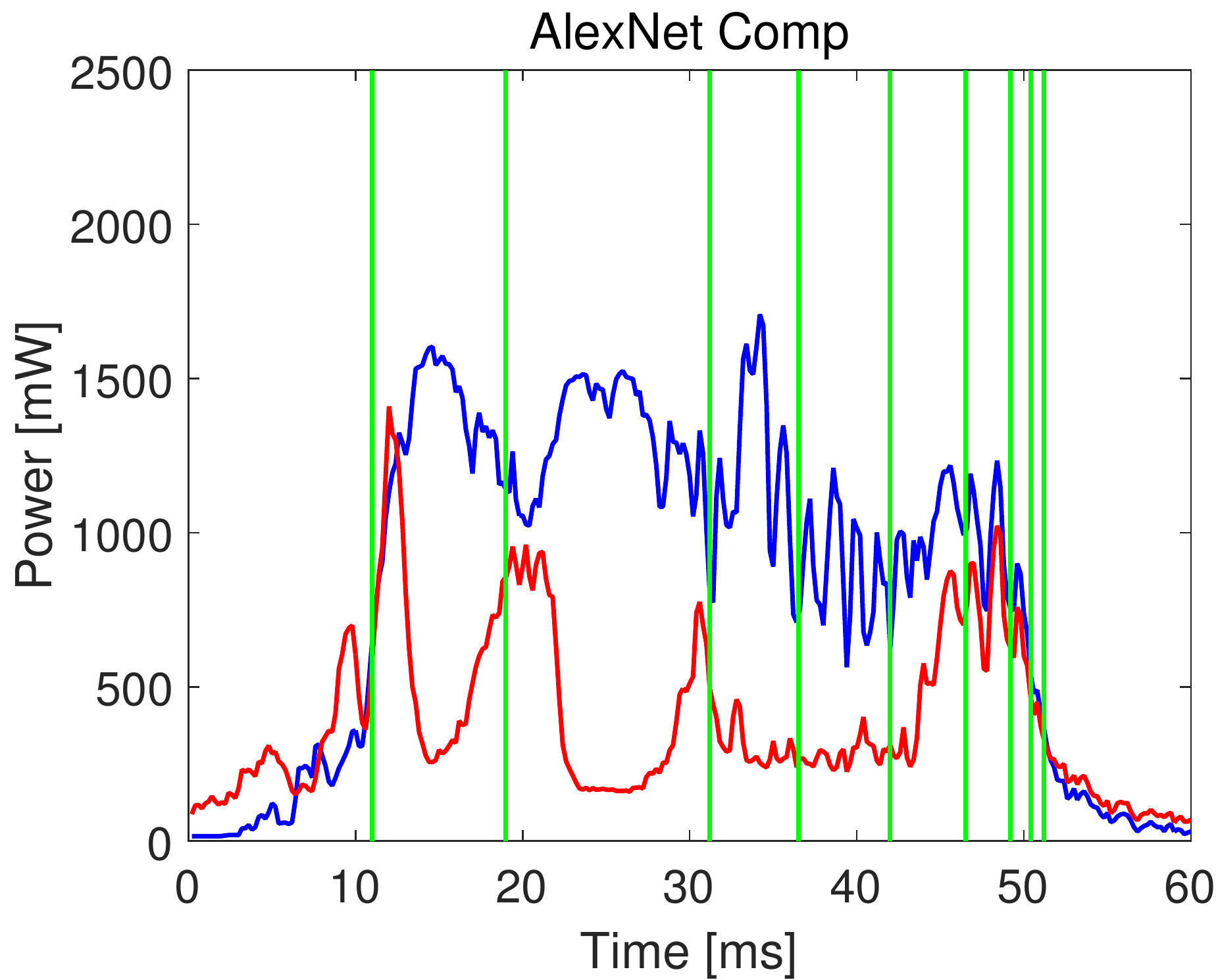}
\includegraphics[width=1.85in]{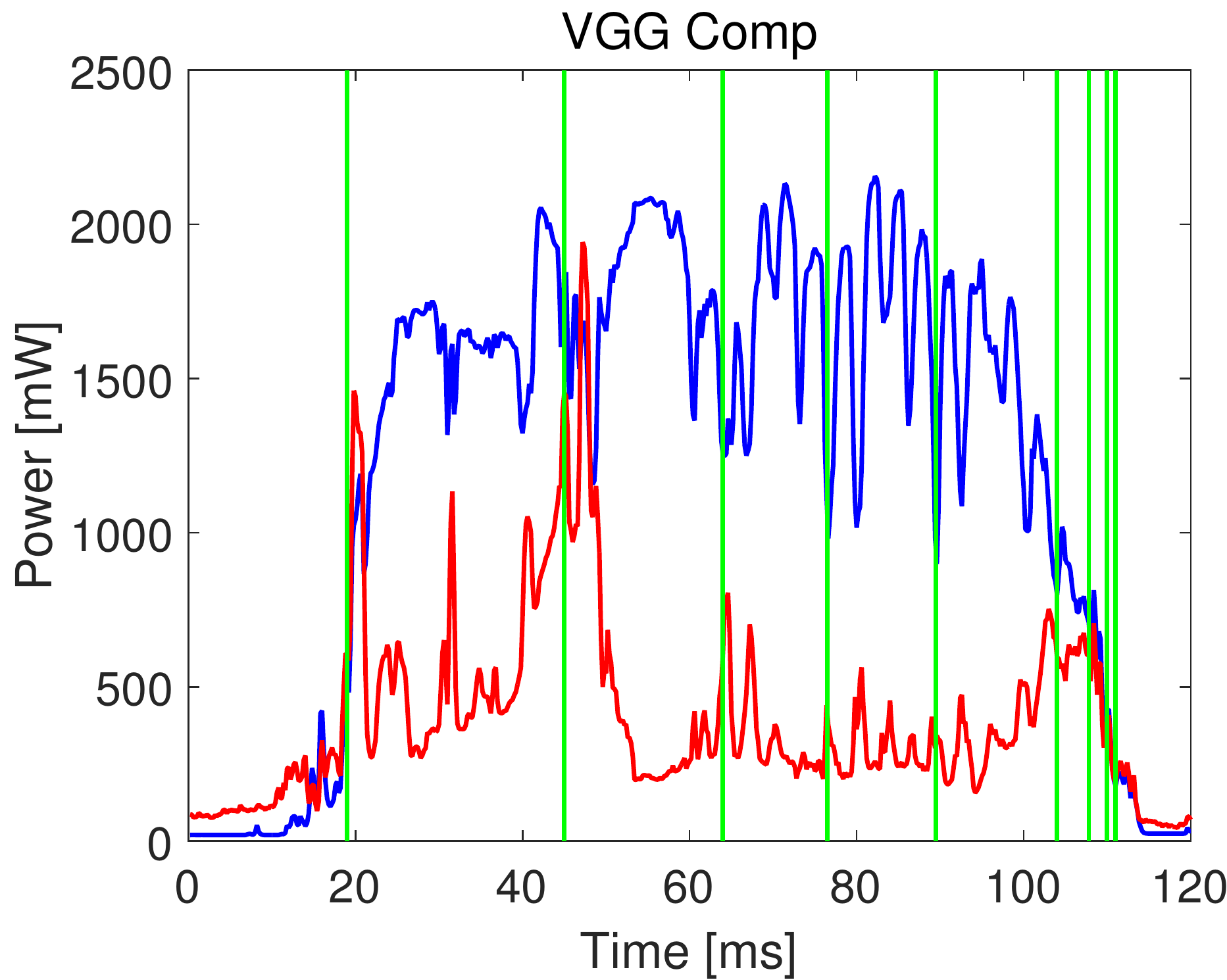}
\includegraphics[width=1.85in]{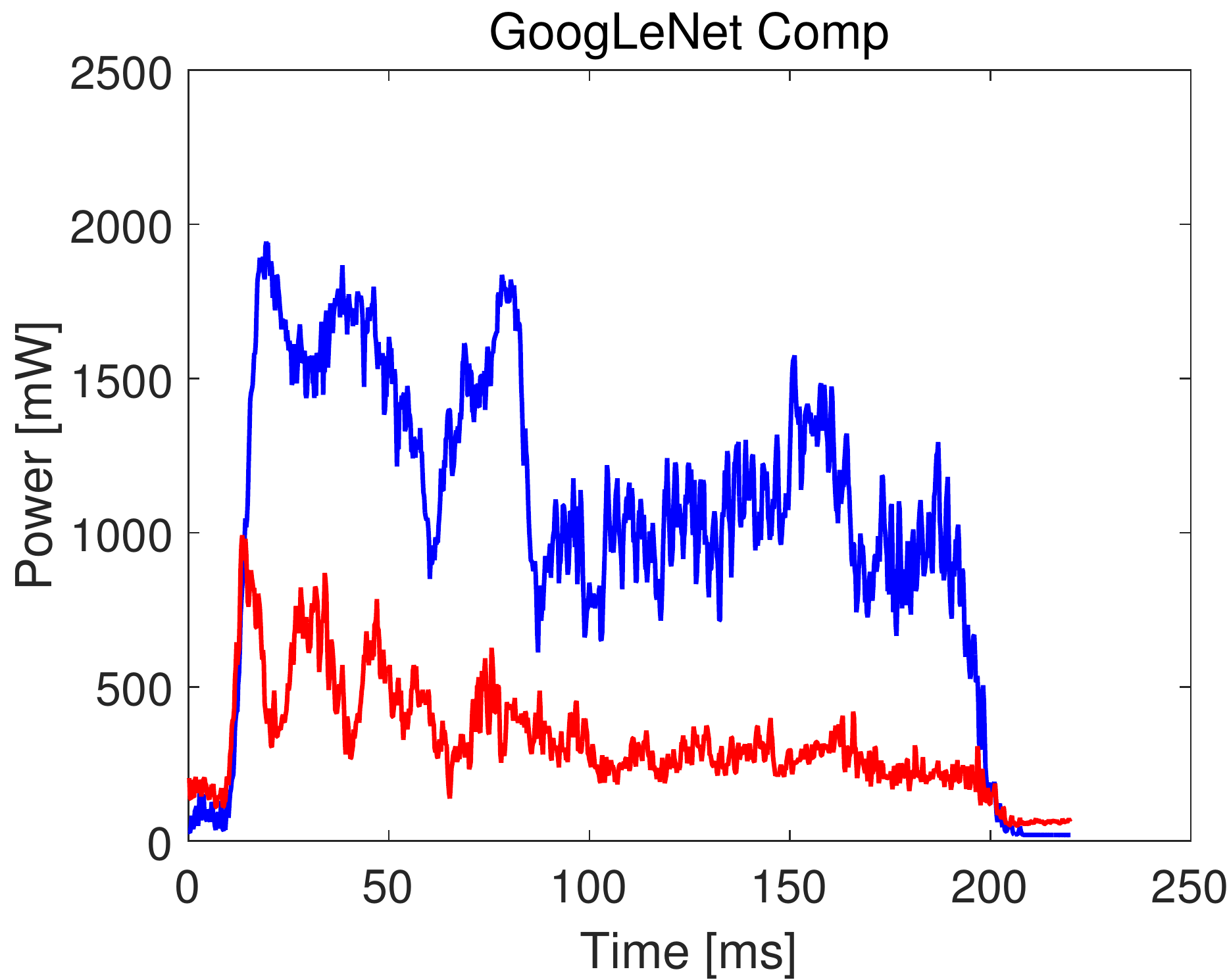}
}

\caption{Power consumption over time for each model. (Blue: GPU, Red: main memory).}
\label{fig:power}
\end{center}
\vskip -0.3in
\end{figure}

\subsection{Energy Consumption Analysis}
Fig. \ref{fig:power} compares power consumption on the smartphone.
Each network gives the power consumption of GPU and main memory. 
Note that we enlarged the time axis of compressed networks for a better comparison. 
We omitted \emph{VGG-16} since \emph{VGG-16} gives similar trend.

The figure shows that the compression reduces 
power consumption (Y axis) as well  as runtime (X axis), 
which explains  why the reduction in energy consumption is larger 
than that in runtime in Table \ref{tab:overall_results}. 
Fig. \ref{fig:power} also shows that the GPU power consumption of compressed CNN 
is smaller than that of uncompressed CNN. We analyze this 
due to the extensive usage of $1\times 1$ convolutions in the compressed CNN. 
When executing convolutions, we apply optimization techniques such as 
\emph{Caffeinated convolution}\citep{chellapilla2006high}. 
In such  a case, in terms of cache efficiency, 
$1\times1$ convolutions are inferior to the other convolutions, 
e.g., $3\times3$, $5\times5$, etc. 
since the amount of data reuse is proportional to the total size of convolution kernel. 
Thus, $1\times1$ convolutions tend to incur more cache misses than the other larger convolutions. 
Cache misses  on the mobile GPU without  sufficient thread level  parallelism often incur stall cycles, 
i.e., make GPU cores idle consuming less power, which reduces the  power consumption of GPU core during 
the execution of $1\times1$ convolution.

As  mentioned earlier, our proposed method improves cache efficiency by reducing the amount of weights. However, $1\times1$ convolutions have negative impacts on cache efficiency and GPU core utilization. 
Fig. \ref{fig:power} shows the combined effects. 
In the compressed networks, the  power consumption of GPU core is reduced by $1\times1$ convolutions 
and tends to change more frequently due to frequent executions of $1\times1$ convolution
while, in the case of uncompressed networks, 
especially for \emph{AlexNet} and \emph{VGG-S}, 
the power consumption of GPU  core tends to be stable during the execution of convolutional layers. 
In the case of uncompressed \emph{GoogLeNet}, the power consumption tends  to fluctuate.
It is mainly because (1) \emph{GoogLeNet} consists of many small  layers (about 100 building blocks), 
and (2) $1\times1$ convolutions are heavily utilized.

The three compressed networks show similar behavior of frequent fluctuations 
in power consumption mostly due to $1\times1$ convolutions.
Fig. \ref{fig:power} also shows that, 
in the uncompressed networks, 
fully connected layers incur significant amount of power consumption in main memory. 
It is because the uncompressed networks, especially  \emph{AlexNet} and \emph{VGG-S} have large numbers 
(more than tens of mega-bytes) of weights in fully connected layers 
which incur significant amount of memory accesses. 
As shown in Fig. \ref{fig:power}, 
the proposed scheme reduces the amount of weights at fully connected layers
thereby reducing the power consumption in main memory.

\section{Discussion}
Although we can obtain very promising results with one-shot rank selection, 
it is not fully investigated yet whether the selected rank is really optimal or not.
As future work, we will investigate the optimality of our proposed scheme. 
The $1\times1$ convolution is a key operation in our compressed model 
as well as in \emph{inception} module of \emph{GoogLeNet}. 
Due to its characteristics, e.g. channel compression and computation reduction, 
we expect that $1\times1$ convolutions will become more and more popular in the future. 
However, as shown in our experimental results, it lacks in cache efficiency. 
We expect further investigations are required to make best use of 1x1 convolutions.

Whole network compression is challenging 
due to the large design space and associated long design time. 
In order to address this problem, 
we propose a one-shot compression scheme 
which applies a single general low-rank approximation method 
and a global rank selection method. 
Our one-shot compression enables fast design 
and easy implementation with publicly available tools.
We evaluated the effectiveness of the proposed scheme on a smartphone and Titan X. 
The experiments show that the proposed scheme gives, 
for four CNNs (\emph{AlexNet}, \emph{VGG-S}, \emph{GoogLeNet}, and \emph{VGG-16})
average $\times2.72$ ($\times3.41$), $\times3.68$ ($\times4.26$), $\times1.42$ ($\times1.60$),
and $\times3.34$ ($\times3.53$)
improvements in runtime (energy consumption) on the smartphone.

\bibliography{iclr2016}
\bibliographystyle{iclr2016_conference}

\section*{Appendices}
\renewcommand{\thesubsection}{\Alph{subsection}}

\subsection{Experimental Setup}
This section describes the details of experimental setup 
including the measurement system for power consumption and exemplifies the measured data.

\subsubsection{Measurement System}
\begin{figure}[h]
\centering
\includegraphics[width=0.9\linewidth]{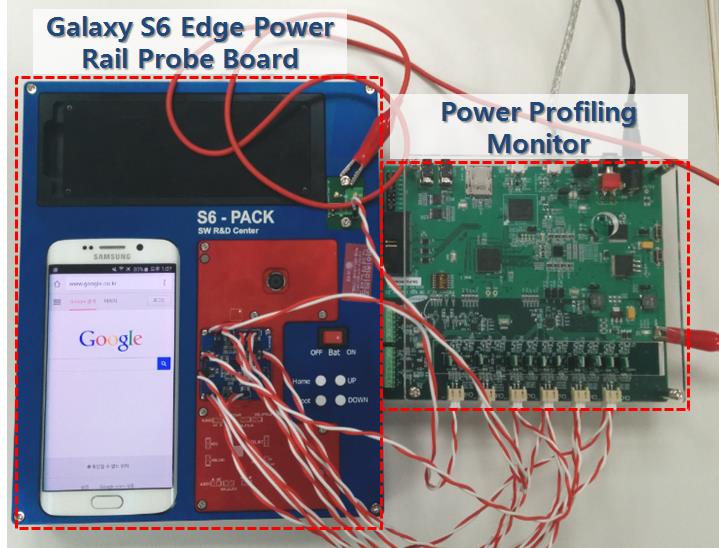}
\caption{Power measurement system.}
\label{fig:power_system}
\end{figure}

Fig. \ref{fig:power_system} shows the power measurement system. 
As the figure shows, it consists of a probe board (left) 
having a Samsung Galaxy S6 smartphone and 
power probes and a monitor board (right). 
The probe board provides 8 probes which are connected to 
the power pins of application processor (to be introduced below). 
The power profiling monitor samples, for each power probe, 
the electric current every 0.1ms and gives power consumption data with time stamps.

\begin{figure}[h]
\centering
\includegraphics[width=0.9\linewidth]{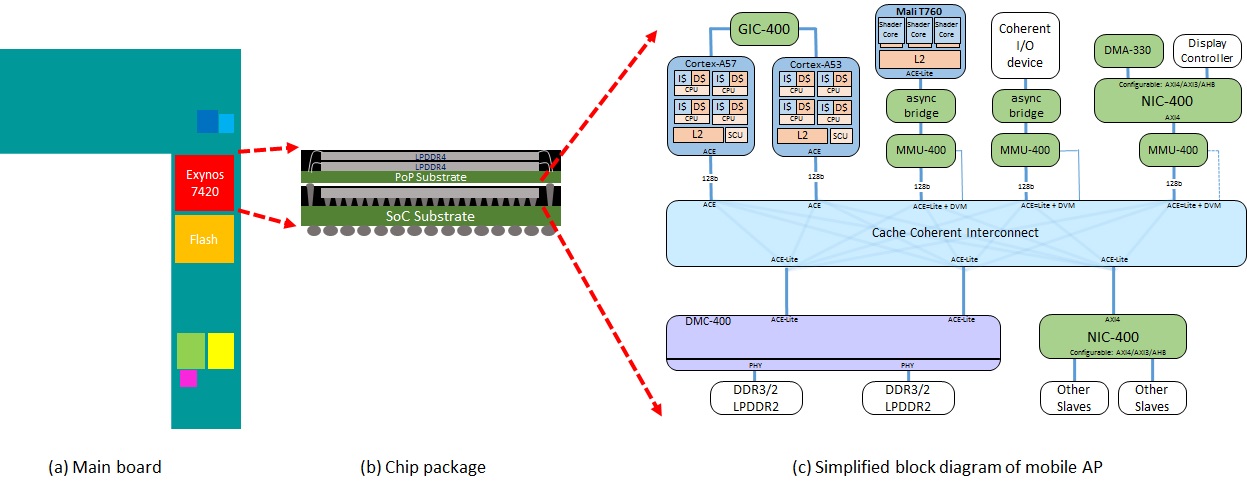}
\caption{Details of mobile application processor and main memory.}
\label{fig:ap}
\end{figure}

Fig. \ref{fig:ap} illustrates the main board of the smartphone (Fig. \ref{fig:ap} (a)), 
the application processor chip package (red rectangle in Fig. 2 (a)) 
consisting of the application processor and main memory (LPDDR4 DRAM) in the smartphone 
(Fig. \ref{fig:ap} (b)), and a simplified block diagram of the application processor 
(Fig. \ref{fig:ap} (c)). 
The power measurement system provides the probes 
connected to the power pins for mobile GPU (ARM Mali T760 in Fig. \ref{fig:ap} (c)) 
and main memory (LPDDR4 DRAM in Fig. \ref{fig:ap} (b)).

\subsubsection{Measured Data Example: \emph{GoogLeNet} Case}
\begin{figure}[h]
\centering
\includegraphics[width=0.9\linewidth]{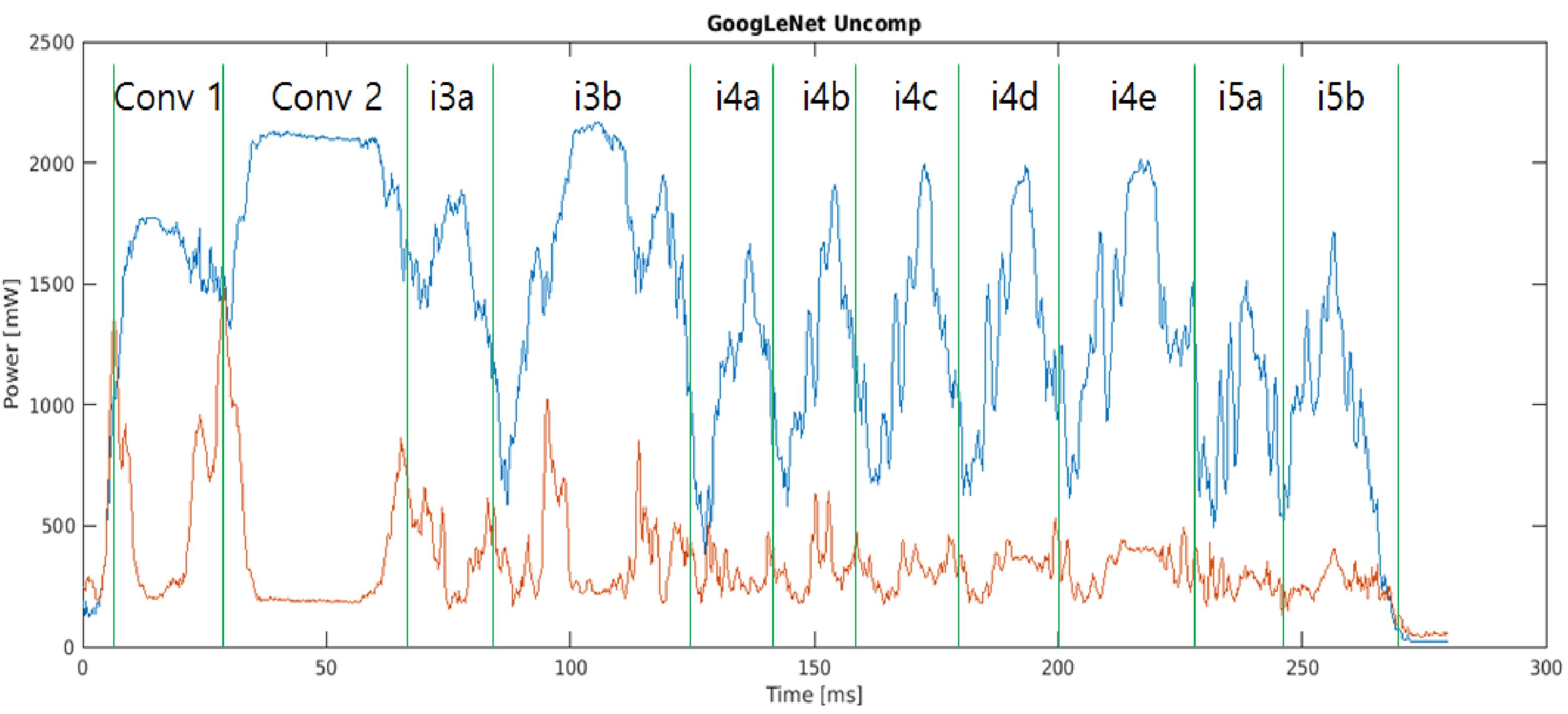}
\caption{Power profile of uncompressed \emph{GoogLeNet}.}
\label{fig:googlenet}
\end{figure}

Fig. \ref{fig:googlenet} shows the power consumption data for the uncompressed GoogLeNet. 
We also identified the period of each layer, 
e.g., the first convolutional layer (Conv 1 in the figure), 
and the first Inception module (i3a). 
As mentioned in our submission, the profile of power consumption 
shows more frequent fluctuations in Inception modules than 
in the convolutional layers. 
The figure also shows that the first two convolutional layers (Conv 1 and Conv 2) 
occupy about 1/4 of total energy consumption 
while Inception modules consume about 3/4 of total energy consumption.

\subsection{Layerwise Analysis}
We report detailed comparison results \emph{VGG-S}, \emph{GoogLeNet}, and \emph{VGG-16}.
\input{table_vgg_s}
\input{table_googlenet}
\input{table_vgg_16}

\end{document}

%% file: table_alexnet.tex
\begin{table*}[t]
\caption{Layerwise analysis on \emph{AlexNet}. 
Note that conv2, conv4, and conv5 layer have 2-group structure.
($S$: input channel dimension, $T$: output channel dimension, $(R_3, R_4)$: Tucker-2 rank).}
\label{tab:AlexNet}
\vskip 0.05in
\begin{center}
\begin{tabular}{l|r|r|r|l|l}
Layer & $S / R3$ & $T / R_4$ & Weights & FLOPs & S6  \\ \hline
conv1 & 3 & 96 & 35K & 105M & 15.05 ms \\
conv1* &  & 26 & 11K & ~~36M(=29+7) & 10.19m(=8.28+1.90)  \\
(imp.)& & & $(\times 2.92)$ & $(\times 2.92)$ &  $(\times 1.48)$  \\ \hline
conv2 & $48\times 2$ & $128 \times 2$  & 307K & 224M & 24.25 ms \\
conv2* & $25 \times 2$ & $59 \times 2$ & ~~91K & 67M(=2+54+11) & 10.53ms(=0.80+7.43+2.30) \\
(imp.)& & & $(\times 3.37)$ & $(\times 3.37)$ & $(\times 2.30)$ \\ \hline
conv3 & 256 & 384  & 885K & 150M & 18.60ms \\
conv3* & 105 & 112 & 178K & ~~30M(=5+18+7) & ~~4.85ms(=1.00+2.72+1.13)\\
(imp.)& & & $(\times 5.03)$ & $(\times 5.03)$ & $(\times 3.84)$  \\ \hline
conv4 & $192\times 2$ & $192\times 2$  & 664K & 112M & 15.17ms \\
conv4* & $49 \times 2$ & $46\times 2$ & ~~77K & ~~13M(=3+7+3) & ~~4.29 
ms(=1.55+1.89+0.86) \\
(imp.)& & & $(\times 7.10)$ & $(\times 7.10)$ & $(\times 3.53)$ \\ \hline
conv5 & $192 \times 2$ & $128 \times 2$  & 442K & 75.0M & 10.78ms\\
conv5* & $40 \times 2$ & $34 \times 2$ & 49K & ~~8.2M(=2.6+4.1+1.5) & 3.44 
ms(=1.15+1.61+0.68) \\
(imp.)& & & $(\times 9.11)$ & $(\times 9.11)$ & $(\times 3.13)$ \\ \hline
fc6 & 256 & 4096  & 37.7M & 37.7M & 18.94ms \\
fc6* & 210 & 584 & 6.9M & ~~8.7M(=1.9+4.4+2.4) & ~~5.07 
ms(=0.85+3.12+1.11)\\
(imp.)& & & $(\times 8.03)$ & $(\times 4.86)$ & $(\times 3.74
)$ \\ \hline
fc7 & 4096 & 4096  & 16.8M & 16.8M & ~~7.75ms \\
fc7* &  & 301 & ~~2.4M &  ~~2.4M(=1.2+1.2) & ~1.02 
~ms(=0.51+0.51) \\
(imp.)& & & $(\times 6.80)$ & $(\times 6.80)$ & $(\times 7.61)$ \\ \hline
fc8 & 4096 & 1000  & ~~4.1M & ~~4.1M & ~~2.00ms \\
fc8* &  & 195 & ~~1.0M &  ~~1.0M(=0.8+0.2) & ~~0.66ms(=0.44+0.22)  \\
(imp.)& & & $(\times 4.12)$ & $(\times 4.12)$ & $(\times 3.01  )$ \\ 
\end{tabular}
\end{center}
\end{table*}

%% file: table_vgg_s.tex
\begin{table*}[h]
\caption{Layerwis analysis on \emph{VGG-S}. 
($S$: input channel dimension, $T$: output channel dimension, $(R_3, R_4)$: Tucker-2 rank).}
\label{tab:VGG_S}
\begin{center}
\begin{tabular}{l|r|r|r|l|l}
Layer & $S / R3$ & $T / R_4$ & Weights & FLOPs & S6 \\ \hline 
conv1 & 3 & 96 & 14K & 168M & 23.88ms \\
conv1* &  & 42 & 10K & 121M(=73+48) & 23.15ms(=14.47+8.68) \\
(imp.)& & & $(\times 1.38)$ & $(\times 1.38)$ &  $(\times 1.03)$\\ \hline
conv2 & 96 & 256  & 614K & 699M & 74.57ms \\
conv2* & 48 & 89 & 134K & 147M(=6+116+25) & 18.54ms(=1.32+13.59+3.64) \\
(imp.)& & & $(\times 4.58)$ & $(\times 4.54)$ & $(\times 4.02)$ \\ \hline
conv3 & 256 & 512  & 1180K & 341M & 38.33ms\\
conv3* & 126 & 175 & 320K & ~~93M(=9+57+26) & 11.59ms(=1.53+6.82+3.24) \\
(imp.)& & & $(\times 3.68)$ & $(\times 3.68)$ & $(\times 3.31)$ \\ \hline
conv4 & 512 & 512  & 2359K & 682M & 78.43ms\\
conv4* & 143 & 144 & 332K & ~~96M(=21+54+21) & 12.23ms(=2.92+6.63+2.78)  \\
(imp.)& & & $(\times 7.10)$ & $(\times 7.10)$ & $(\times 6.37)$ \\ \hline
conv5 & 512 & 512  & 2359K & 682M & 78.40ms \\
conv5* & 120 & 120 & 252K& ~~73M(=18+37+18) & 10.25ms(=2.76+5.03+2.46)\\
(imp.)& & & $(\times 9.34)$ & $(\times 9.34)$ & $(\times 7.65)$  \\ \hline
fc6 & 512 & 4096  & 75.5M & 75.5M & 40.75ms \\
fc6* & 343 & 561 & 9.4M & 15.5M(=6.3+6.9+2.3) & ~~7.18ms(=1.58+4.61+0.98) \\
(imp.)& & & $(\times 8.03)$ & $(\times 4.86)$ & $(\times 5.68)$  \\ \hline
fc7 & 4096 & 4096  & 16.8M & 16.8M & ~~7.68ms \\
fc7* &  & 301 & 2.4M &  ~~2.4M(=1.2+1.2) & ~~1.26ms(=0.65+0.60)\\
(imp.)& & & $(\times 6.80)$ & $(\times 6.80)$ & $(\times 6.10 )$ \\ \hline
fc8 & 4096 & 1000  & ~~4.1M & ~~4.1M & ~~1.97ms \\
fc8* &  & 195 & 1.0M &  ~~1.0M(=0.8+0.2) & ~~0.67ms(=0.45+0.22) \\
(imp.)& & & $(\times 4.12)$ & $(\times 4.12)$ & $(\times 2.92 )$ \\ 
\end{tabular}
\end{center}
\end{table*}

%% file: table_googlenet.tex
\begin{table*}[h]
\caption{Layerwise analysis on \emph{GoogLeNet}. 
($S$: input channel dimension, $T$: output channel dimension, $(R_3, R_4)$: Tucker-2 rank).}
\label{tab:GoogLeNet}
\vskip 0.05in
\begin{center}
\begin{tabular}{l|r|r|r|l|l}
Layer & $S / R3$ & $T / R_4$ & Weights & FLOPs & S6  \\ \hline 
conv1 & 3 & 64 & 9.4K & 118M & 18.96ms \\
conv1* &  & 23 & 4.8K & ~~60M(=42+18) &21.76ms(=16.85+4.91) \\
(imp.)& & & $(\times 1.94)$ & $(\times 1.94)$ &  $(\times 0.87)$ \\ \hline
conv2 & 64 & 192 & 11.1K & 347M & 34.69ms \\
conv2* &  & 23 & 4.8K & ~~60M(=42+18) & 12.04ms(=1.66+5.95+4.43)\\
(imp.)& & & $(\times 4.99)$ & $(\times 4.99)$ &  $(\times 2.88)$ \\ \hline
i3a & 96 & 128 & 111K$(68\%)$ & 87M$(68\%)$ & 9.39ms \\
i3a* & 41 & 41 & 24K & 19M(3+12+4) & 	3.70ms=(0.70+2.02+0.98)\\
(imp.)& & & $(\times 4.55)$ & $(\times 4.55)$ &  $(\times 2.54)$ \\ \hline
i3b & 128 & 192 & 221K$(57\%)$ & 173M$(57\%)$ & 17.49ms  \\
i3b* & 42 & 37 & 26K & ~~21M(4+11+6) & 	4.10ms=(0.89+1.99+1.21)\\
(imp.)& & & $(\times 8.36 )$ & $(\times 8.36 )$ &  $(\times 4.27 )$  \\ \hline
i4a & 96 & 208  & 180K$(48\%)$ & 35M$(48\%)$ & 4.35ms \\
i4a* & 35 & 39  & 24K & ~~5M(1+2+2) & 1.68ms=(0.39+0.79+0.50)  \\
(imp.)& & & $(\times 7.56 )$ & $(\times7.56 )$ &  $(\times 2.60 )$  \\ \hline
i4b & 112 & 224 & 226K$(51\%)$ & 44M$(51\%)$ & 5.39ms  \\
i4b* & 55 & 75 & 60K & 12M(1+7+3) & 2.65ms=(0.47+1.48+0.70) \\
(imp.)& & & $(\times 3.76 )$ & $(\times 3.76)$ &  $(\times 2.03)$  \\ \hline
i4c & 128 & 256 & 295K$(58\%)$ & 58M$(58\%)$ & 6.93ms  \\
i4c* & 63 & 87 & 80K & 16M(2+10+4) & 3.10ms=(0.52+1.74+0.84)  \\
(imp.)& & & $(\times 3.70)$ & $(\times 3.70)$ &  $(\times 2.23)$ \\ \hline
i4d & 144  & 288 & 373K$(62\%)$ & 73M$(62\%)$ & 8.93ms  \\
i4d* &  67 & 105 & 103K & 20M(2+12+6) & 3.67ms=(0.61+2.03+1.04)  \\
(imp.)& & & $(\times 3.62)$ & $(\times 3.62)$ &  $(\times 2.43)$  \\ \hline
i4e & 160 & 320  & 461K$(60\%)$ & 90M$(60\%)$ & 10.90ms \\
i4e* & 97 & 131 & 172K & 34M(3+22+8) & 5.45ms=(0.76+3.35+1.34)  \\
(imp.)& & & $(\times 2.68 )$ & $(\times 2.68 )$ &  $(\times 2.00)$  \\ \hline
i5a & 160 & 320 & 461K$(44\%)$ & 23M$(44\%)$ & 3.96ms  \\
i5a* & 91 & 139 & 173K & ~~8M(1+6+2) & 2.55ms=(0.41+1.55+0.59) \\
(imp.)&  &  & $(\times  2.67)$ & $(\times 2.67)$ &  $(\times 1.55)$ \\ \hline
i5b & 192 & 384 & 664K$(46\%)$ & 33M$(46\%)$ & 5.71ms  \\
i5b* & 108 & 178  & 262K ~~~ & 13M(1+8+3) ~~~~~~~~~~~~~~~ & 3.28ms=(0.51+1.95+0.82) ~~~ \\
(imp.)& ~~~~~~~~ & ~~~~~~ & $(\times 2.53)$ & $(\times 2.53)$ &  $(\times 1.74)$ \\ 
\end{tabular}
\end{center}
\end{table*}

%% file: table_vgg_16.tex
\begin{table*}[h]
\caption{Layerwis analysis on \emph{VGG-16}. We do not compress the first convolutional layer
and fully-connected layers as done in \cite{zhang2015b}.
The theoretical speed-up raito of convolutional layers and whole layers are
$\times5.03$ and $\times 4.93$ respectively.
($S$: input channel dimension, $T$: output channel dimension, $(R_3, R_4)$: Tucker-2 rank).}
\label{tab:VGG_16}
\begin{center}
\begin{tabular}{l|r|r|r|l|l}
Layer & $S / R3$ & $T / R_4$ & Weights & FLOPs & S6 \\ \hline 
$C1_2$ & 64 & 64 & 37K & 1853M & 234.58ms \\
$C1_2$* & 11 & 18 & 4K & ~~182M(=35+89+58) & 64.35ms(= 13.42+28.46+20.47) \\
(imp.)& & & $(\times 10.15)$ & $(\times 10.15)$ &  $(\times 3.65)$\\ \hline
$C2_1$ & 64 & 128  & 74K & 926M & 105.66ms \\
$C2_1$* & 22 & 34 & 8K & 101M(=8+38+55) & 23.20ms(=3.26+7.82+12.12) \\
(imp.)& & & $(\times 9.17)$ & $(\times 9.17)$ & $(\times 4.55)$ \\ \hline
$C2_2$ & 128 & 128  & 148K & 1851M & 226.29ms\\
$C2_2$* & 39 & 36 & 22K & ~~279(=63+159+58) & 50.66ms(=11.60+27.19+11.86) \\
(imp.)& & & $(\times 6.64)$ & $(\times 6.64)$ & $(\times 4.47)$ \\ \hline
$C3_1$ & 128 & 256  & 295K & 926M & 93.71ms\\
$C3_1$* & 58 & 117 & 74K & ~~231M(=15+122+94) & 29.92ms(=2.96+14.73+12.23)  \\
(imp.)& & & $(\times 4.01)$ & $(\times 4.01)$ & $(\times 3.13)$ \\ \hline
$C3_2$ & 2562 & 256  & 590K & 1850M & 211.75ms \\
$C3_2$* & 138 & 132 & 76K& ~~237M(=55+129+53) & 34.16ms(=7.94+17.89+8.33)\\
(imp.)& & & $(\times 7.81)$ & $(\times 7.81)$ & $(\times 6.20)$  \\ \hline
$C3_3$ & 256 & 256  & 590K & 1850M & 213.31ms \\
$C3_3$* & 124 & 119 & 195K& ~~612M(=100+416+96) & 72.66ms(=12.74+47.74+12.19)\\
(imp.)& & & $(\times 3.03)$ & $(\times 3.03)$ & $(\times 2.94)$  \\ \hline
$C4_1$ & 256 & 512  & 1180K & 925M & 98.40ms \\
$C4_1$* & 148 & 194 & 265K& ~~208M(=17+114+78) & 23.58ms(=2.54+12.25+8.79)\\
(imp.)& & & $(\times 4.45)$ & $(\times 4.45)$ & $(\times 4.17)$  \\ \hline
$C4_2$ & 512 & 512  & 2360K & 1850M & 216.16ms \\
$C4_2$* & 212 & 207 & 609K& ~~478M(=85+310+83) & 51.18ms(=9.10+33.22+8.86)\\
(imp.)& & & $(\times 3.87)$ & $(\times 3.87)$ & $(\times 4.22)$  \\ \hline
$C4_3$ & 512 & 512  & 2360K & 1850M & 216.34ms \\
$C4_3$* & 178 & 163 & 436K& ~~342M(=71+205+65) & 38.85ms(=8.06+23.14+7.66)\\
(imp.)& & & $(\times 5.42)$ & $(\times 5.42)$ & $(\times 5.57)$  \\ \hline
$C5_1$ & 512 & 512  & 2360K & 463M & 57.54ms \\
$C5_1$* & 185 & 164 & 452K& ~~89M(=19+54+16) & 13.09ms(=2.80+7.89+2.39)\\
(imp.)& & & $(\times 5.22)$ & $(\times 5.22)$ & $(\times 4.40)$  \\ \hline
$C5_2$ & 512 & 512  & 2360K & 463M & 76.80ms \\
$C5_2$* & 172 & 170 & 416K& ~~82M(=16+48+17) & 11.87ms(=2.64+6.82+2.42)\\
(imp.)& & & $(\times 5.67)$ & $(\times 5.67)$ & $(\times 6.47)$  \\ \hline
$C5_3$ & 512 & 512  & 2360K & 463M & 67.69ms \\
$C5_3$* & 120 & 120 & 438K& ~~86M(=17+52+17) & 12.16ms(=2.65+7.13+2.38)\\
(imp.)& & & $(\times 5.38)$ & $(\times 5.38)$ & $(\times 5.57)$  \\ \hline
\end{tabular}
\end{center}
\end{table*}

%% file: KimYD_iclr2016.bbl
\begin{thebibliography}{38}
\providecommand{\natexlab}[1]{#1}
\providecommand{\url}[1]{\texttt{#1}}
\expandafter\ifx\csname urlstyle\endcsname\relax
  \providecommand{\doi}[1]{doi: #1}\else
  \providecommand{\doi}{doi: \begingroup \urlstyle{rm}\Url}\fi

\bibitem[Bader et~al.(2015)Bader, Kolda, et~al.]{TTB_Software}
Bader, Brett~W., Kolda, Tamara~G., et~al.
\newblock Matlab tensor toolbox version 2.6.
\newblock Available online, February 2015.
\newblock URL \url{http://www.sandia.gov/~tgkolda/TensorToolbox/}.

\bibitem[Carroll \& Chang(1970)Carroll and Chang]{carroll1970}
Carroll, J~Douglas and Chang, Jih-Jie.
\newblock Analysis of individual differences in multidimensional scaling via an
  n-way generalization of “eckart-young” decomposition.
\newblock \emph{Psychometrika}, 35\penalty0 (3):\penalty0 283--319, 1970.

\bibitem[Chellapilla et~al.(2006)Chellapilla, Puri, and
  Simard]{chellapilla2006high}
Chellapilla, Kumar, Puri, Sidd, and Simard, Patrice.
\newblock High performance convolutional neural networks for document
  processing.
\newblock In \emph{Tenth International Workshop on Frontiers in Handwriting
  Recognition}. Suvisoft, 2006.

\bibitem[Chen et~al.(2015)Chen, Wilson, Tyree, Weinberger, and Chen]{chen2015}
Chen, Wenlin, Wilson, James~T, Tyree, Stephen, Weinberger, Kilian~Q, and Chen,
  Yixin.
\newblock Compressing neural networks with the hashing trick.
\newblock \emph{arXiv preprint arXiv:1504.04788}, 2015.

\bibitem[Cheng et~al.(2015)Cheng, Yu, Feris, Kumar, Choudhary, and
  Chang]{cheng2015}
Cheng, Yu, Yu, Felix~X, Feris, Rogerio~S, Kumar, Sanjiv, Choudhary, Alok, and
  Chang, Shih-Fu.
\newblock Fast neural networks with circulant projections.
\newblock \emph{arXiv preprint arXiv:1502.03436}, 2015.

\bibitem[De~Lathauwer et~al.(2000)De~Lathauwer, De~Moor, and
  Vandewalle]{de2000}
De~Lathauwer, Lieven, De~Moor, Bart, and Vandewalle, Joos.
\newblock A multilinear singular value decomposition.
\newblock \emph{SIAM journal on Matrix Analysis and Applications}, 21\penalty0
  (4):\penalty0 1253--1278, 2000.

\bibitem[De~Silva \& Lim(2008)De~Silva and Lim]{de2008}
De~Silva, Vin and Lim, Lek-Heng.
\newblock Tensor rank and the ill-posedness of the best low-rank approximation
  problem.
\newblock \emph{SIAM Journal on Matrix Analysis and Applications}, 30\penalty0
  (3):\penalty0 1084--1127, 2008.

\bibitem[Denil et~al.(2013)Denil, Shakibi, Dinh, de~Freitas, et~al.]{denil2013}
Denil, Misha, Shakibi, Babak, Dinh, Laurent, de~Freitas, Nando, et~al.
\newblock Predicting parameters in deep learning.
\newblock In \emph{Advances in Neural Information Processing Systems}, pp.\
  2148--2156, 2013.

\bibitem[Denton et~al.(2014)Denton, Zaremba, Bruna, LeCun, and
  Fergus]{denton2014}
Denton, Emily~L, Zaremba, Wojciech, Bruna, Joan, LeCun, Yann, and Fergus, Rob.
\newblock Exploiting linear structure within convolutional networks for
  efficient evaluation.
\newblock In \emph{Advances in Neural Information Processing Systems}, pp.\
  1269--1277, 2014.

\bibitem[Glorot \& Bengio(2010)Glorot and Bengio]{glorot2010}
Glorot, Xavier and Bengio, Yoshua.
\newblock Understanding the difficulty of training deep feedforward neural
  networks.
\newblock In \emph{International conference on artificial intelligence and
  statistics}, pp.\  249--256, 2010.

\bibitem[Gong et~al.(2014)Gong, Liu, Yang, and Bourdev]{gong2014}
Gong, Yunchao, Liu, Liu, Yang, Ming, and Bourdev, Lubomir.
\newblock Compressing deep convolutional networks using vector quantization.
\newblock \emph{arXiv preprint arXiv:1412.6115}, 2014.

\bibitem[Han et~al.(2015{\natexlab{a}})Han, Mao, and Dally]{han2015b}
Han, Song, Mao, Huizi, and Dally, William~J.
\newblock A deep neural network compression pipeline: Pruning, quantization,
  huffman encoding.
\newblock \emph{arXiv preprint arXiv:1510.00149}, 2015{\natexlab{a}}.

\bibitem[Han et~al.(2015{\natexlab{b}})Han, Pool, Tran, and Dally]{han2015a}
Han, Song, Pool, Jeff, Tran, John, and Dally, William~J.
\newblock Learning both weights and connections for efficient neural networks.
\newblock \emph{arXiv preprint arXiv:1506.02626}, 2015{\natexlab{b}}.

\bibitem[Harshman \& Lundy(1994)Harshman and Lundy]{harshman1994}
Harshman, Richard~A and Lundy, Margaret~E.
\newblock Parafac: Parallel factor analysis.
\newblock \emph{Computational Statistics \& Data Analysis}, 18\penalty0
  (1):\penalty0 39--72, 1994.

\bibitem[He et~al.(2015)He, Zhang, Ren, and Sun]{he2015}
He, Kaiming, Zhang, Xiangyu, Ren, Shaoqing, and Sun, Jian.
\newblock Delving deep into rectifiers: Surpassing human-level performance on
  imagenet classification.
\newblock In \emph{IEEE International Conference on Computer Vision}, 2015.

\bibitem[Hinton et~al.(2012)Hinton, Srivastava, Krizhevsky, Sutskever, and
  Salakhutdinov]{hinton2012}
Hinton, Geoffrey~E, Srivastava, Nitish, Krizhevsky, Alex, Sutskever, Ilya, and
  Salakhutdinov, Ruslan~R.
\newblock Improving neural networks by preventing co-adaptation of feature
  detectors.
\newblock \emph{arXiv preprint arXiv:1207.0580}, 2012.

\bibitem[Ioffe \& Szegedy(2015)Ioffe and Szegedy]{ioffe2015}
Ioffe, Sergey and Szegedy, Christian.
\newblock Batch normalization: Accelerating deep network training by reducing
  internal covariate shift.
\newblock In \emph{International Conference on Machine Learning}, 2015.

\bibitem[Jaderberg et~al.(2014)Jaderberg, Vedaldi, and
  Zisserman]{jaderberg2014}
Jaderberg, M., Vedaldi, A., and Zisserman, A.
\newblock Speeding up convolutional neural networks with low rank expansions.
\newblock In \emph{British Machine Vision Conference}, 2014.

\bibitem[Kim \& Choi(2007)Kim and Choi]{kim2007}
Kim, Y.-D. and Choi, S.
\newblock Nonnegative {Tucker} decomposition.
\newblock In \emph{Proceedings of the IEEE CVPR-2007 Workshop on Component
  Analysis Methods}, Minneapolis, Minnesota, 2007.

\bibitem[Kolda \& Bader(2009)Kolda and Bader]{kolda2009}
Kolda, Tamara~G and Bader, Brett~W.
\newblock Tensor decompositions and applications.
\newblock \emph{SIAM review}, 51\penalty0 (3):\penalty0 455--500, 2009.

\bibitem[Lebedev et~al.(2015)Lebedev, Ganin, Rakhuba, Oseledets, and
  Lempitsky]{lebedev2015}
Lebedev, Vadim, Ganin, Yaroslav, Rakhuba, Maksim, Oseledets, Ivan, and
  Lempitsky, Victor.
\newblock Speeding-up convolutional neural networks using fine-tuned
  cp-decomposition.
\newblock In \emph{International Conference on Learning Representations}, 2015.

\bibitem[Lin et~al.(2014)Lin, Chen, and Yan]{lin2014}
Lin, M., Chen, Q., and Yan, S.
\newblock Network in network.
\newblock In \emph{International Conference on Learning Representations}, 2014.

\bibitem[MacKay(1992)]{mackay1992}
MacKay, David~JC.
\newblock Bayesian interpolation.
\newblock \emph{Neural computation}, 4\penalty0 (3):\penalty0 415--447, 1992.

\bibitem[Mathieu et~al.(2013)Mathieu, Henaff, and LeCun]{mathieu2013}
Mathieu, Michael, Henaff, Mikael, and LeCun, Yann.
\newblock Fast training of convolutional networks through ffts.
\newblock \emph{arXiv preprint arXiv:1312.5851}, 2013.

\bibitem[M{\o}rup \& Hansen(2009)M{\o}rup and Hansen]{morup2009}
M{\o}rup, Morten and Hansen, Lars~Kai.
\newblock Automatic relevance determination for multi-way models.
\newblock \emph{Journal of Chemometrics}, 23\penalty0 (7-8):\penalty0 352--363,
  2009.

\bibitem[Nakajima(2015)]{nakajima2015}
Nakajima, Shinichi.
\newblock \emph{Variational Bayesian matrix factorization version 1.02}, 2015.
\newblock URL \url{https://sites.google.com/site/shinnkj23/downloads}.

\bibitem[Nakajima et~al.(2012)Nakajima, Tomioka, Sugiyama, and
  Babacan]{nakajima2012}
Nakajima, Shinichi, Tomioka, Ryota, Sugiyama, Masashi, and Babacan, S~Derin.
\newblock Perfect dimensionality recovery by variational bayesian pca.
\newblock In \emph{Advances in Neural Information Processing Systems}, pp.\
  971--979, 2012.

\bibitem[Nakajima et~al.(2013)Nakajima, Sugiyama, Babacan, and
  Tomioka]{nakajima2013}
Nakajima, Shinichi, Sugiyama, Masashi, Babacan, S~Derin, and Tomioka, Ryota.
\newblock Global analytic solution of fully-observed variational bayesian
  matrix factorization.
\newblock \emph{The Journal of Machine Learning Research}, 14\penalty0
  (1):\penalty0 1--37, 2013.

\bibitem[Novikov et~al.(2015)Novikov, Podoprikhin, Osokin, and
  Vetrov]{novikov2015}
Novikov, Alexander, Podoprikhin, Dmitry, Osokin, Anton, and Vetrov, Dmitry.
\newblock Tensorizing neural networks.
\newblock \emph{arXiv preprint arXiv:1509.06569}, 2015.

\bibitem[Shashua \& Hazan(2005)Shashua and Hazan]{shashua2005}
Shashua, Amnon and Hazan, Tamir.
\newblock Non-negative tensor factorization with applications to statistics and
  computer vision.
\newblock In \emph{Proceedings of the 22nd international conference on Machine
  learning}, pp.\  792--799. ACM, 2005.

\bibitem[Simonyan \& Zisserman(2015)Simonyan and Zisserman]{simonyan2015}
Simonyan, K. and Zisserman, A.
\newblock Very deep convolutional networks for large-scale image recognition.
\newblock In \emph{International Conference on Learning Representations}, 2015.

\bibitem[Szegedy et~al.(2015)Szegedy, Liu, Jia, Sermanet, Reed, Anguelov,
  Erhan, Vanhoucke, and Rabinovich]{szegedy2015}
Szegedy, Christian, Liu, Wei, Jia, Yangqing, Sermanet, Pierre, Reed, Scott,
  Anguelov, Dragomir, Erhan, Dumitru, Vanhoucke, Vincent, and Rabinovich,
  Andrew.
\newblock Going deeper with convolutions.
\newblock In \emph{IEEE Conference on Computer Vision and Pattern Recognition
  (CVPR)}, 2015.

\bibitem[Tipping(2001)]{tipping2001}
Tipping, Michael~E.
\newblock Sparse bayesian learning and the relevance vector machine.
\newblock \emph{The journal of machine learning research}, 1:\penalty0
  211--244, 2001.

\bibitem[Tucker(1966)]{tucker1966}
Tucker, Ledyard~R.
\newblock Some mathematical notes on three-mode factor analysis.
\newblock \emph{Psychometrika}, 31\penalty0 (3):\penalty0 279--311, 1966.

\bibitem[Vanhoucke et~al.(2011)Vanhoucke, Senior, and Mao]{vanhoucke2011}
Vanhoucke, Vincent, Senior, Andrew, and Mao, Mark~Z.
\newblock Improving the speed of neural networks on cpus.
\newblock In \emph{Proc. Deep Learning and Unsupervised Feature Learning NIPS
  Workshop}, volume~1, 2011.

\bibitem[Ye(2005)]{ye2005}
Ye, Jieping.
\newblock Generalized low rank approximations of matrices.
\newblock \emph{Machine Learning}, 61\penalty0 (1-3):\penalty0 167--191, 2005.

\bibitem[Zhang et~al.(2015{\natexlab{a}})Zhang, Zou, He, and Sun]{zhang2015b}
Zhang, Xiangyu, Zou, Jianhua, He, Kaiming, and Sun, Jian.
\newblock Accelerating very deep convolutional networks for classification and
  detection.
\newblock \emph{arXiv preprint arXiv:1505.06798}, 2015{\natexlab{a}}.

\bibitem[Zhang et~al.(2015{\natexlab{b}})Zhang, Zou, Ming, He, and
  Sun]{zhang2015a}
Zhang, Xiangyu, Zou, Jianhua, Ming, Xiang, He, Kaiming, and Sun, Jian.
\newblock Efficient and accurate approximations of nonlinear convolutional
  networks.
\newblock 2015{\natexlab{b}}.

\end{thebibliography}
